\RequirePackage{amsmath}
\documentclass[runningheads]{llncs}

\newcommand*\mybar[1]{%
\scriptsize
  \hbox{%
    \vbox{%
      \hrule height 0.6pt % The actual bar
      \kern0.2ex%         % Distance between bar and symbol
      \hbox{%
        \kern -0.2em%      % Shortening on the left side
        \ensuremath{{#1}}%
        \kern 0.1em%      % Shortening on the right side
      }%
    }%
  }%
}

\newcommand*\bs[1]{%
\boldsymbol{#1}
}

\newcommand*{\longeq}{\scalebox{2}[1]{=}
}

\usepackage{color}
\newcommand{\ignore}[1]{}

\usepackage{amsmath}
\usepackage{amssymb}
\usepackage{mathtools}
\usepackage{amsfonts}
\usepackage{lipsum}
\usepackage{hyperref}
\usepackage{booktabs}
\usepackage{bm,subfigure,epsfig}
\usepackage{array}
\usepackage{mdwtab}
\usepackage{multirow}
\usepackage[latin1]{inputenc}
\usepackage{algorithm}
\usepackage{algpseudocode}

\usepackage{cite}
\usepackage{graphicx}

\ifpdf
  \DeclareGraphicsExtensions{.eps,.pdf,.png,.jpg}
\else
  \DeclareGraphicsExtensions{.eps}
\fi

\usepackage{amsopn}

\newcounter{relctr} %% <- counter for relations
\everydisplay\expandafter{\the\everydisplay\setcounter{relctr}{0}} %% <- reset every eq
\AtBeginDocument{} %% <- store original definition

\newfam\openfam

\def\qed{\hfill$\square$}

\ifpdf
\hypersetup{
  pdftitle={Fast OMP for Exact Recovery and Sparse Approximation},
  pdfauthor={Huiyuan Yu, Jia He, Maggie Cheng}
}
\fi

\begin{document}

\title{Fast Orthogonal Matching Pursuit through Successive Regression}

\author{ 
Huiyuan Yu \orcidID{0000-0003-1052-5612} 
\and Jia He 
\and Maggie Cheng \orcidID{0000-0003-1823-4043}
}

\authorrunning{H. Yu et al.}
% First names are abbreviated in the running head.
% If there are more than two authors, 'et al.' is used.
%
\institute{Illinois Institute of Technology, Chicago, IL, 60616\\
\email{hyu47@hawk.iit.edu, jhe58@hawk.iit.edu, maggie.cheng@iit.edu}}

\maketitle
\begin{abstract}

Orthogonal Matching Pursuit (OMP) has been a powerful method in sparse signal recovery and approximation. However, OMP suffers computational issues when the signal has a large number of non-zeros. This paper advances OMP and its extension called generalized OMP (gOMP) by offering fast algorithms for the orthogonal projection of the input signal at each iteration. The proposed modifications directly reduce the computational complexity of OMP and gOMP. Experiment results verified the improvement in computation time. This paper also provides sufficient conditions for exact signal recovery. For general signals with additive noise, the approximation error is at the same order as OMP (gOMP), but is obtained within much less time.

\keywords{Greedy Algorithm \and Compressive Sensing \and Sparse Signal Recovery \and Approximation \and Orthogonal Matching Pursuit
}
\end{abstract}

\section{Introduction}

Let $\bs x$ be a $d$-dimensional real signal. Suppose there is a real measurement matrix $\Phi \in \mathbb{R}^{N \times d}$, through which we can obtain an $N$-dimensional measurement  $ \bs y= \Phi \bs x $. Usually $N < d$, which presents an underdetermined system. How to reconstruct the original signal $\bs x$ from an underdetermined system? If $\bs x$ is sparse, then by exploiting sparsity, we may be able to find a unique solution. $\bs x$  is called a $k$-sparse signal if $\bs x$ has at most $k$ non-zero components.

The measurement matrix $\Phi$ is also called a dictionary, and each column $\bs \varphi$ of the dictionary called an atom. Let $\mathcal{J}=\{1, \ldots, d\}$ represent the index set of all atoms in the dictionary. If the dictionary is overcomplete, there are many representations of $ \bs y= \sum\limits_{\gamma  \in \mathcal{J} } a_\gamma \bs \varphi_\gamma$. Intuitively, we would like to find the sparsest solution:  $\min\limits_{\bs x} \|\bs x \|_0  \textrm{ subject to } \bs y= \Phi \bs x$,  but it is an NP-hard problem. Different optimization principles lead to different sparse representations of $\bs y$, for example, basis pursuit (BP) \cite{BP-PNAS, BP, BP-short} and the method of frames (MOF) \cite{MOF} among many others \cite{weighted, difference}:

\begin{itemize}
\item Find a representation of the input signal whose coefficients have the minimal $\ell_1$ norm. 
\begin{equation} \tag{BP}
\min_{\bs x} \|\bs x \|_1 \;\; \textrm{ subject to }\;  \bs y= \Phi \bs x 
\end{equation}
\item Find a representation of the  input signal whose coefficients have the minimal $\ell_2$ norm.
\begin{equation} \tag{MOF}
\min_{\bs x} \|\bs x \|_2 \;\; \textrm{  subject to } \; \bs y= \Phi \bs x 
\end{equation}
\end{itemize}
BP and MOF both provide convex relaxation to the $\ell_0$ norm minimization problem, however, neither of them provides the sparsest solution, except those satisfying the sparsity condition specified in \cite{BP-sparse}.

Matching Pursuit (MP) \cite{MP-first} uses an iterative procedure that directly addresses the sparsity issue. 
Orthogonal Matching Pursuit (OMP) \cite{OMP-first, tropp2004greed, tropp2007signal} inherits the greedy approach from MP that selects an atom with the maximal correlation with the residual at present, but improves over the standard MP by adding least square minimization at each iteration. Let $\Gamma$ be the index set of atoms found so far, the least square estimation is used for computing the orthogonal projection of the input signal $\bs y$ onto the subspace spanned by the atoms indexed by $\Gamma$:

\begin{equation} \tag{OMP}
\min_{\bs x_\Gamma} \| \bs y- \Phi_\Gamma \bs x_\Gamma \|_2^2 \;\; \text{ with } |\Gamma| \le k
\end{equation}

OMP has been shown to have better results than MP. Many variations of OMP have been developed \cite{sOMP, multipathMP, SP, rOMP, needell2009cosamp, gOMP}. Under certain conditions OMP provides recovery guarantee \cite{RIP, tropp2007signal, tropp2004greed, RIPGap,candes2008restricted, RIP-OMP-improved}. The excellent performance  of OMP results from the orthogonal projection of $\bs y$ onto the subspace spanned by the atoms selected so far.  The least square solution is obtained by $\bs x_\Gamma = \Phi_\Gamma^+ \bs y$. As $\Gamma$ increases, solving the least square problem significantly increases the computational load. In this paper, we propose a fundamental improvement over classical OMP to avoid the high complexity of computing pseudo inverse over an increasing-sized matrix, which can be generalized to other OMP-based algorithms: 

\begin{itemize}
\item When solving the least square problem at each iteration, instead of computing $\Phi_\Gamma^+ \bs y$ over the entire support $\Gamma$, it uses successive regression over a single atom. It makes the same greedy choice as OMP does at each iteration, but is much faster due to reduced computation load. The  proposed algorithm is called OMP-SR.

\item The blocked version of OMP is called Generalized Orthogonal Matching Pursuit \cite{gOMP}, which extends the greedy choice to multiple atoms at each iteration but still preserve the convergence property of OMP. We propose a blocked version of OMP-SR, called Blocked Successive Regression (BSR). BSR is an improvement over gOMP, analagous to OMP-SR being an improvement over OMP.
\end{itemize}

In general, the measurement $\bs y$ is often with noise. A general signal may be represented as the linear combination of atoms from the dictionary with additive noise,
\[\bs y = \Phi {\bs x}+ \bs \varepsilon . \]

We are interested in the best approximation of $\bs y$ using a linear combination of atoms. {\it The best approximation} is the one with the smallest approximation error measured by the $\ell_2$ norm of the residual, and hence, the optimization principle is,

\begin{equation} \tag{Sparse Approximation}
\min_{\bs x}  \| \bs y -\Phi \bs x \|_2 \;\; \textrm{   subject to }  \|\bs x\|_0   \le \!  k  
\end{equation}

OMP is a fundational approach for signal reconstruction, therefore, any direct improvement over OMP can benefit many applications that use various implementation of OMP. The proposed method is also different from previous efforts that use matrix factorization based solutions (e.g., \cite{QR-GS,QR1}) and matrix inversion bypass (MIB) technique (\cite{MIB, MIB-app}). In \cite{OMP-comparison}, a simulation-based comparison have been provided over various implementation of OMP. In this paper we not only provide simulation based comparison, but also analytical complexity analysis.

The rest of the paper is organzied as follows: In Section \ref{sec:algo}, we present our algorithms for exact recover; in Section \ref{sec:theory}, we show the main theoretical results for the BSR algorithm\footnote{The proofs can be found in the Appendix.}; in Section \ref{sec:experiment}, we show the performance of our algorithms in real datasets compared to the baseline methods OMP and gOMP. 

\subsection{Notation}

\begin{itemize}
\item $\Phi^\top$: transpose of matrix $\Phi$

\item $\Phi^+$: pseudo inverse of matrix $\Phi$

\item $(\Phi^\top \Phi)^{-1}$: inverse of  matrix $(\Phi^\top \Phi)$

\item $\|A\|_{p \rightarrow q} =\sup\limits_{{\bs x} \ne 0} \frac{\|A {\bs x}\|_q}{\|{\bs x}\|_p}$: operator norm of matrix $A$. 

\item $\|A\|_{p \rightarrow p} $ is abbreviated to $\|A\|_p$ .
\end{itemize}

\section{Recovery Algorithms by Successive Regression}
\label{sec:algo}

\subsection{Orthogonal Matching Pursuit through Successive Regression (OMP-SR)} 

OMP-SR is a fast implementation of OMP. When solving the least square problem at each iteration of OMP, it avoids the expensive computation for the pseudo inverse of $\Phi_{J^{t-1}}$; instead, it only projects onto the atom selected in the current iteration via univariate regression, and then updates the coefficients of atoms selected in previous iterations through a backtracking procedure: $b_l= \beta_l -  \sum\limits_{k=l+1}^t b_k \gamma_{l, k} $(see Algorithm \ref{alg:omp-sr}), where $\beta_t = \frac{ \langle \boldsymbol{z}_t, \boldsymbol{y} \rangle }{ \langle \boldsymbol{z}_t, \boldsymbol{z}_t \rangle} $  is the coefficient newly obtained in the current iteration, $b_l$ is the updated coefficient for the atoms selected in previous iterations. Note that the inner product $\langle \boldsymbol{z}_l, \boldsymbol{z}_l \rangle$ does not need to be recomputed. It only needs to be computed once, that is when we compute $\beta_l$ in the $l$-th iteration.

\begin{algorithm}[!tbh]
   \caption{OMP-SR}
   \label{alg:omp-sr}
\begin{algorithmic}
\State {\bfseries Initialization: }$\boldsymbol{a}_0 = \boldsymbol{z}_0 =\boldsymbol{1}$, $\boldsymbol{r}^0= \boldsymbol{y}$, $J^{0}=\phi$
\For {$t=1$ {\bfseries to} $\kappa$}
\State Choose $j^t=\mathrm{arg}\max\limits_{j \in \mathcal{J} \setminus J^{t-1}} \left|  \langle \bs \varphi_{j}, \boldsymbol{r}^{t-1} \rangle \right|$
\Comment $\bs \varphi_j$ is the $j$-th column of $\Phi$
\State Let $\boldsymbol{a}_t$ be the $j^t$-th column of matrix $\Phi$.
  \State Regress $\boldsymbol{a}_t$ on $\boldsymbol{z}_l$ and get coefficients
  \[ \gamma_{l, t} = \frac{ \langle \boldsymbol{z}_l, \boldsymbol{a}_t \rangle }{ \langle \boldsymbol{z}_l, \boldsymbol{z}_l \rangle}, \;   \text{ for } l=0, \dots, t-1 \]
  \State Compute $\boldsymbol{z}_t= \boldsymbol{a}_t - \sum\limits_{l=0}^{t-1} \gamma_{l,t} \boldsymbol{z}_l$
  \State Regress $\boldsymbol{y}$ on $\boldsymbol{z}_t$ to get 
  $ \beta_t = \frac{ \langle \boldsymbol{z}_t, \boldsymbol{y} \rangle }{ \langle \boldsymbol{z}_t, \boldsymbol{z}_t \rangle} $
  \State Let $b_t=\beta_t$
\If {$t > 1$}  
 \For{$l= t-1$ {\bfseries to} $1$}
  \State $b_l= \beta_l -  \sum\limits_{k=l+1}^t b_k \gamma_{l, k}$
 \EndFor
\EndIf
  \State Update index set $J^{t}=J^{t-1} \bigcup \{j^t\}$
  \State Update residual $\boldsymbol{r}^t =\boldsymbol{y} -  \sum\limits_{l=1}^t b_l \bs \varphi_{j^l}$
\EndFor
\State Let $\boldsymbol{x}_{j^t} = b_t$ for $t=1, \ldots, \kappa$, and let $\boldsymbol{x}_j = 0$ for $j \notin J^\kappa$
\State Return $\boldsymbol{x}$
\end{algorithmic}
\end{algorithm}

OMP-SR selects the same atom and generates the same residual as OMP does at each iteration, and therefore returns the same result as OMP. OMP-SR starts to show performance gain over OMP when the number of non-zeros in $\bs x$ increases due to not having to compute the pseudo inverse of a growing matrix.

\subsection{Complexity Comparison with QR-based OMP}

In practice, OMP implementation based on incremental QR decomposition may be used for improved efficiency (e.g., \cite{QR-GS, QR1, QR2, QR3, QR4}). In each iteration, $Q_t$ and $R_t$ matrices are updated as in the algorithm. To obtain the updated solution for the least square problem, one needs to compute $ {\bs h}=Q_t^\top  {\bs y}$, and then use back-substitution to solve $R_t {\bs x}={\bs h}$. However, despite the cost saving over standard OMP, the operation cost of OMP based on QR decomposition is still higher than that of the proposed OMP-SR. Table \ref{tbl:flops-qr} and Table \ref{tbl:flops-sr} show the floating-point operations of them for each iteration of the OMP algorithm.

\begin{table}[!htbp]
\centering
\caption{Operation cost for the $t$-th iteration of OMP using QR update}
\label{tbl:flops-qr}
\begin{tabular}{c | c}
\hline
Operation & Flops \\
\hline
Update $Q_t$, $R_t$ & $(d-t)(4N-1) + 3N + 1$ \\
Update $ {\bs h} = Q_t^\top {\bs y}$ &  $2N$ \\
Solve $R_t {\bs x} = {\bs h} $ & $t^2$ \\
Total cost & $(d-t)(4N-1)+5N+1+t^2 $\\
\hline
\end{tabular}
\end{table}

\begin{table}[!htbp]
\centering
\caption{Operation cost for the $t$-th iteration of the proposed OMP-SR}
\label{tbl:flops-sr}
\begin{tabular}{c | c}
\hline
Operation & Flops \\
\hline
Compute $\gamma_{l,t}$ & $2Nt$ \\
Compute $\beta_t$ &   $4N-1$ \\
Update coefficients $b_l$ & $t^2-t$ \\
Total cost  & $t(2N-1)+4N-1+t^2$\\
\hline
\end{tabular}
\end{table}

A term-by-term comparison shows OMP-SR uses fewer flops than QR-based OMP. The cost analysis is for $\Phi  \in \mathbb{R}^{N \times d}$. For sparse signals with $k$ non-zeros, as long as \; $t(2N-1)<(d-t)(4N-1)$, OMP-SR outperforms QR-based OMP by a margin of at least $N+2$ per iteration. Typically in OMP, $t \le k \ll d$ for sparse recovery problems, therefore, the condition $t(2N-1)<(d-t)(4N-1)$ is easily satisfied.

\subsection{Blocked Successive Regression (BSR)}
BSR builds on the idea of successive regression in OMP-SR but selects a block of atoms at each iteration. The block size $c$ is a hyper parameter, usually decided by a grid search. The algorithm is still greedy in nature: in each iteration it selects the atoms that have the largest correlations with the residual measured by the $\ell_2$ norm. Each iteration of BSR performs an orthogonal projection of $\bs y$ over $c$ newly selected atoms, instead of over $|\Gamma|$ atoms, which could be costly as $|\Gamma |$ increases with iterations. Subsequently the coefficients for atoms selected in previous iterations are updated  through  $ {b}_{i}= {\beta}_i -  \sum\limits_{k = l+1}^t \sum\limits_{j \in \Gamma_k} {b}_{j} \gamma_{i, j}$ (see Algorithm \ref{alg:bsr}).

\begin{algorithm}[!tbh]
   \caption{BSR}
   \label{alg:bsr}
\begin{algorithmic}
\State {\bfseries Initialization:} $\boldsymbol{r}^0= \boldsymbol{y}$, $\Gamma = \phi$, $\boldsymbol{z}_0 =\boldsymbol{1}$
\For {$t=1$ {\bfseries to} $\kappa$}
\State $\Gamma_t=\mathrm{arg}\max\limits_{\begin{subarray}{c}{\Omega:|\Omega|=c} \\{ \Omega \subset \mathcal{J}\setminus \Gamma} \end{subarray} } \left\|  \Phi^\top_{\Omega} \boldsymbol{r}^{t-1} \right\|_2$

\For {each $j \in \Gamma_t$}
 \State Let $\boldsymbol{a}_j$ be the $j$-th column of matrix $\Phi$.
 \State Regress $\boldsymbol{a}_j$ on $\boldsymbol{z}_0$ to get coefficient
   $ \gamma_{0, j} = \frac{ \langle \boldsymbol{z}_0, \boldsymbol{a}_j \rangle }{ \langle \boldsymbol{z}_0, \boldsymbol{z}_0 \rangle} $
 \State Compute $\boldsymbol{z}_j= \boldsymbol{a}_j - \gamma_{0,j} \boldsymbol{z}_0$
 \If { $t >1 $ }
\State Regress $\boldsymbol{a}_j$ on $Z_{\Gamma_l}$ to get coefficients 
  \[\boldsymbol{\gamma}_{\Gamma_l, j} = Z_{\Gamma_l}^+ \boldsymbol{a}_j, \; \text{for } l=1, \ldots, t-1 \]
\State Compute $\boldsymbol{z}_j= \boldsymbol{z}_j - \sum\limits_{l=1}^{t-1} \sum\limits_{i \in \Gamma_l} \gamma_{i, j} \boldsymbol{z}_i$
\EndIf
\EndFor
\State Regress $\boldsymbol{y}$ on $Z_{\Gamma_t}$ to get coefficients
  $ \boldsymbol{\beta}_{\Gamma_t}=Z_{\Gamma_t}^+ \boldsymbol{y} $
\State Let $ \boldsymbol{b}_{\Gamma_t}= \boldsymbol{\beta}_{\Gamma_t}$   

\If {$t > 1$}  
\For{$l=t-1$ {\bfseries to} $1$}
\For{$i \in \Gamma_l$}
  \State ${b}_{i}= {\beta}_i -  \sum\limits_{k = l+1}^t \sum\limits_{j \in \Gamma_k} {b}_{j} \gamma_{i, j}$
 \EndFor
 \EndFor
\EndIf
\State Update index set $\Gamma=\Gamma \bigcup \Gamma_t$
\State Update residual $\boldsymbol{r}^t =\boldsymbol{y} -  \Phi_{\Gamma} \boldsymbol{b}_{\Gamma} $
\State Break if $\|\boldsymbol{r}^t\|_2 \le \delta$
\EndFor
\State Let $\boldsymbol{x}_{\Gamma} = \boldsymbol{b}_{\Gamma}$, and let $\boldsymbol{x}_{\mathcal{J} \setminus \Gamma} = 0$
\State Return $\boldsymbol{x}$

\end{algorithmic}
\end{algorithm}

The BSR algorithm halts if the residual becomes too small or it has exhausted $\kappa$ iterations, which amounts to two of the three halting rules listed in \cite{needell2009cosamp} for matching pursuit type of algorithms.

The columns selected by BSR shall be the same as the columns selected by gOMP \cite{gOMP} in each iteration. However, the two algorithms differ in the way they solve the least square problem.
\section{Conditions for Exact Recovery}
\label{sec:theory}
\subsection{Background}

Assume there are $k$ non-zero entries in a $d$-dimensional signal $\bs x$, and $k \ll d$. Let $\Lambda_\text{opt}=\{i_1, \ldots, i_k\}$ be the set of indices for the non-zero entries of  $\bs x$. Without loss of generality, we can partition the measurement matrix as $ \Phi=[\Phi_\text{opt} | \Psi]$ so that $\Phi_\text{opt}$ has $k$ columns, $\Phi_\text{opt}=[\bs \varphi_{i_1}, \ldots, \bs \varphi_{i_k}]$, and  $\Psi$ has the remaining $d-k$ columns.

In the absence of noise, the measured signal $\bs y$ has a sparse representation: $\bs y = \Phi \bs x = \sum\limits_{j \in \Lambda_\mathrm{opt}} a_j \bs \varphi_j$. Exact recovery aims to recover the coefficients $a_j$ for all atoms indexed by $\Lambda_\mathrm{opt}$, which are the non-zero entries in $\bs x$.

\subsubsection{The Exact Recovery Condition of OMP-SR}
Algorithm OMP-SR essentially is an OMP algorithm with fast implementation: it starts with the same initial residual $\boldsymbol{r}^0$ and selects the same atom in the next iteration, so the residual $\boldsymbol{r}^t$ after the $t$-th iteration is the same. Since $\boldsymbol{r}^t$ is used as input to the next iteration when choosing a column, the next iteration will result in the same residual $\boldsymbol{r}^{t+1}$. By induction, after $k$ iterations the algorithm returns the same result as OMP does. The exact recovery condition for OMP-SR is the same as for OMP.

\subsection{The Exact Recovery Conditions of BSR}

BSR is essentially a greedy algorithm, which makes a greedy choice at each iteration, except that BSR selects a block of columns at each iteration with a fixed block size $c$ ($c \ge 1$). If $c=1$, BSR reduces to OMP-SR. We have learned that under the condition of $\rho(\bs r) <1$, OMP and OMP-SR can find one optimal column in each iteration. Then for BSR, under what condition will each iteration of BSR only select the optimal columns from $\Phi_\mathrm{opt}$ except the last iteration? This is the best case, in which BSR can locate all optimal columns within $\lceil k/c \rceil$ iterations. We call the condition for the best case as {\it the strong exact recovery condition for BSR}.

\vspace{0.1in}

\subsubsection{A Strong Exact Recovery Condition for BSR}
Recall that $\Phi=[\Phi_\text{opt} | \Psi]$ so that $\Phi_\text{opt}$ has the $k$ optimal columns, and  $\Psi$ has the remaining $d-k$ columns. Let $\boldsymbol{r}$ denote the residual at the current iteration before the greedy choice is made. 

\vspace{0.1in}

For a fixed block size $c$, the greedy choice ratio is defined as follows:
\begin{equation}
\label{eq:rho-c-def}
\rho_c(\bs r) \stackrel{\text { def }}{\longeq} \frac{\max\limits_{\Omega_1}\left\| \Phi^\top_{\Omega_1} \boldsymbol{r}\right\|_{2}}{\max\limits_{\Omega_2}\left\|  \Phi^\top_{\Omega_2}\boldsymbol{r}\right\|_{2}},
\end{equation}
such that $|\Omega_1|=|\Omega_2|=c$ and $|\Omega_2 \cap \Lambda_\mathrm{opt} | > |\Omega_1 \cap \Lambda_\mathrm{opt} | $, i.e., $\Omega_2$ has at least one more optimal column than $\Omega_1$. Given a $k$-sparse signal, BSR can recover the signal within $\lceil k/c \rceil$ iterations if the following condition holds.

\begin{theorem}[The strong exact recovery condition for BSR]
\label{thm:strong}
A sufficient condition for BSR to recover a $k$-sparse signal within $\lceil k/c\rceil $ iterations is that
\begin{equation}
\label{eq:rho-c}
\rho_c(\boldsymbol{r}) < 1
\end{equation}
holds for all iterations.
\end{theorem}

\subsubsection{A Weak Exact Recovery Condition for BSR}

What is the condition for $\rho_c(\bs r) <1$ to hold in Theorem \ref{thm:strong}? In the absence of a straightforward answer, we first discuss the condition for BSR to recover a $k$-sparse signal within $k$ iterations, then revisit the condition \eqref{eq:rho-c}.

We call the condition for BSR to recover a $k$-sparse signal within $k$ iterations {\it the weak exact recovery condition for BSR}. For the weak condition, we use the following greedy choice ratio: $\rho(\boldsymbol{r}) \stackrel{}{=} \frac{\left\|{\Psi}^\top \boldsymbol{r}\right\|_{\infty}}{\left\|\Phi_{\mathrm{opt}}^\top \boldsymbol{r}\right\|_{\infty}}$.

\begin{theorem}[The Weak Exact Recovery Condition for BSR]
\label{thm:weak}
A sufficient condition for BSR to recover a $k$-sparse signal within $k$ iterations is that
\begin{equation}
\label{eq:rho-bsr}
\rho(\boldsymbol{r}) < 1
\end{equation}
holds for all iterations.
\end{theorem}

Although intuitive, the condition in \eqref{eq:rho-bsr} expressed in terms of the greedy choice ratio cannot be checked before we know the residuals in all iterations. We need to establish a sufficient condition for the exact recovery by BSR in terms of the property of the dictionary $\Phi$.

BSR will select at least one optimal column at each iteration but can also possibly select some non-optimal columns. We can split the columns of $\Psi$ into two parts: $\Psi=[\Psi_J | \Psi_{\mybar{J}}]$, where $\Psi_J $ are the non-optimal columns that have been selected by BSR algorithm so far, and $\Psi_{\mybar{J}}$ include the remaining columns.
\vspace{0.1in}

Let matrix $X$ be the submatrix of $\Phi$ that includes all columns of $\Phi_{\text{opt}}$ and the columns in $\Psi$ that have been selected by BSR at the previous iterations, i.e., $ X=[\Phi_\text{opt} | \Psi_J] $.
\vspace{0.1in}

Let $\Pi$ denote the index set for the columns in $\Phi_\mathrm{opt}$ that have not been selected by the algorithm so far,  so $| \Pi | \le k$. Let $( \cdot )_\Pi$ denote the columns in the matrix indexed by $\Pi$, and $( \cdot )_{\Pi,:}$ denotes the rows of the matrix indexed by $\Pi$.

\begin{lemma} 
\label{eq:lm-rho}
If  $\max\limits_{\boldsymbol{\psi}} \left\| \left(X^{+} \right)_{\Pi,:} \boldsymbol{\psi}\right\|_1 <1$, where vector $\boldsymbol{\psi}$ ranges over columns of $\Psi_{\mybar{J}}$, then the residual $\boldsymbol{r}$ satisfies $\rho(\boldsymbol{r})<1$.
\end{lemma}

Although condition $\max\limits_{\boldsymbol{\psi}\in \Psi_{\mybar{J}}} \left\| \left( X^{+} \right)_{\Pi,:} \boldsymbol{\psi}\right\|_1 <1$ is expressed in terms of the property of the dictionary, this condition still cannot be checked without executing the algorithm. In practice it is unlikely that the optimal columns are known {\it a priori}, so the submatrices $X, \Psi_{\mybar{J}}$ cannot be located before the execution of the algorithm. More practical methods are needed to check the sufficient condition without the execution of the algorithm.

In \cite{tropp2004greed}, a fundamental property of the dictionary $\Phi$, called  {\it coherence} is defined as:

\begin{equation}
\label{coherence}
\mu \overset{\text{def}}{=} \max\limits_{j \ne k} |\langle  \boldsymbol{\varphi}_j,\boldsymbol{\varphi}_k  \rangle|
\end{equation}

Coherence $\mu$ is the maximum absolute value of pairwise inner product between the columns of the dictionary.

For a positive integer $m$, the cumulative coherence function, $\mu_1(m)$ of the dictionary, is defined as
\begin{equation}
\label{eq:mu}
\mu_1(m) \overset{\text{def}}{=}\max\limits_{|\Lambda|=m}\max\limits_{\boldsymbol{\psi}} \sum\limits_{j \in \Lambda} | \langle \boldsymbol{\varphi}_j, \boldsymbol{\psi} \rangle|
\end{equation}

where $\Lambda$ is the set of indices for any $m$ columns of $\Phi$, and $\boldsymbol{\psi}$ ranges over the columns of $\Phi$ not indexed by $\Lambda$. $\mu_1(m)$ is the maximum cumulative coherence from any $m$ columns of $\Phi$.

Next, we use the cumulative coherence property of the dictionary to derive a sufficient condition.
\begin{lemma}
\label{eq:lm-mu}
$\max\limits_{\boldsymbol{\psi}\in \Psi_{\mybar{J}}} \left\| \left(X^{+}\right)_{\Pi,:} \boldsymbol{\psi}\right\|_1 <1$ whenever $\mu_1(l)+\mu_1(n)<1$ holds, where $n$ is the number of columns in $X$, and $l =\min ( | \Pi |, k-1 )$.
\end{lemma}

\vspace{10pt}
 
Lemma \ref{eq:lm-mu} and Lemma \ref{eq:lm-rho} together lead to the following conclusion: the residual $\bs r$ satisfies $\rho(\bs r)<1$ whenever 
\begin{equation}
\label{eq:mu1-sum}
\mu_1(l)+ \mu_1(n)<1.
\end{equation}

\subsubsection{Revisit Theorem \ref{thm:strong}: the Sufficient Condition for $\rho_c({\bf r})<1$.}

It is easy to show that $\mu_1(l)+ \mu_1(n)<1$ is also sufficient for $\rho_c({\bs r}) < 1$ to hold in Theorem \ref{thm:strong}, which leads to the following theorem.

\begin{theorem}[The strong exact recovery condition for BSR]
\label{eq:thm-strong-mu}
Suppose that $\mu$ is the coherence of the dictionary as defined in \eqref{coherence}. A sufficient condition for BSR to recover a $k$-sparse signal within $\lceil k/c \rceil$ iterations is that
\begin{equation}
\label{eq:ultimate}
\mu(2k-1)<1.
\end{equation}
\end{theorem}

\section{Experiments}
\label{sec:experiment}
 
Data used in the experiments are posted at \href{https://github.com/arsarting/Compressed-Sensing}{github}.

\begin{figure}[!htbp]
\centering
\begin{tabular}{cccc}
\subfigure[]{\epsfig{file=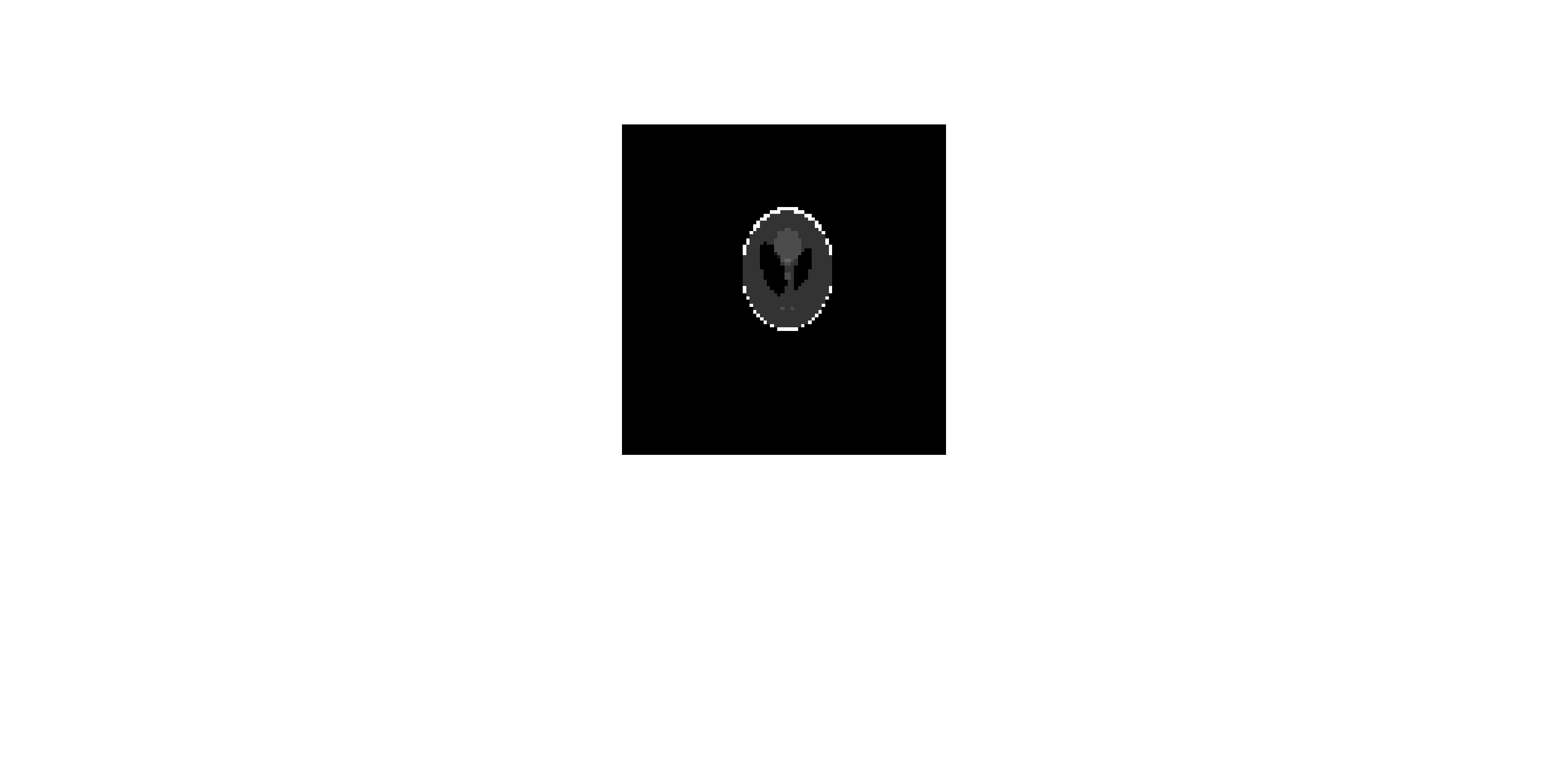, width=0.34\textwidth, bb= 500 300 900 600, clip=}}&
\subfigure[]{\epsfig{file=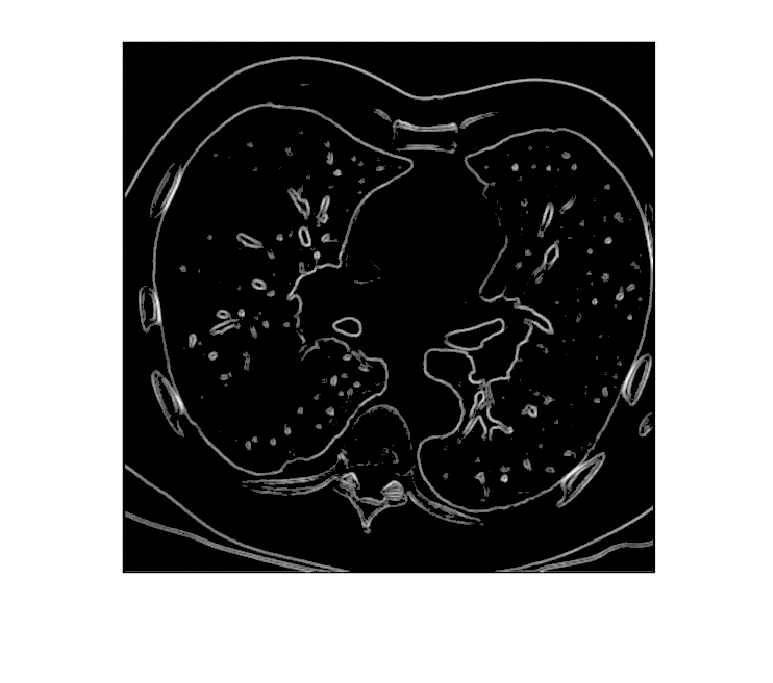, width=0.34\textwidth, bb= 50 50 380 320, clip=}}\\
\subfigure[]{\epsfig{file=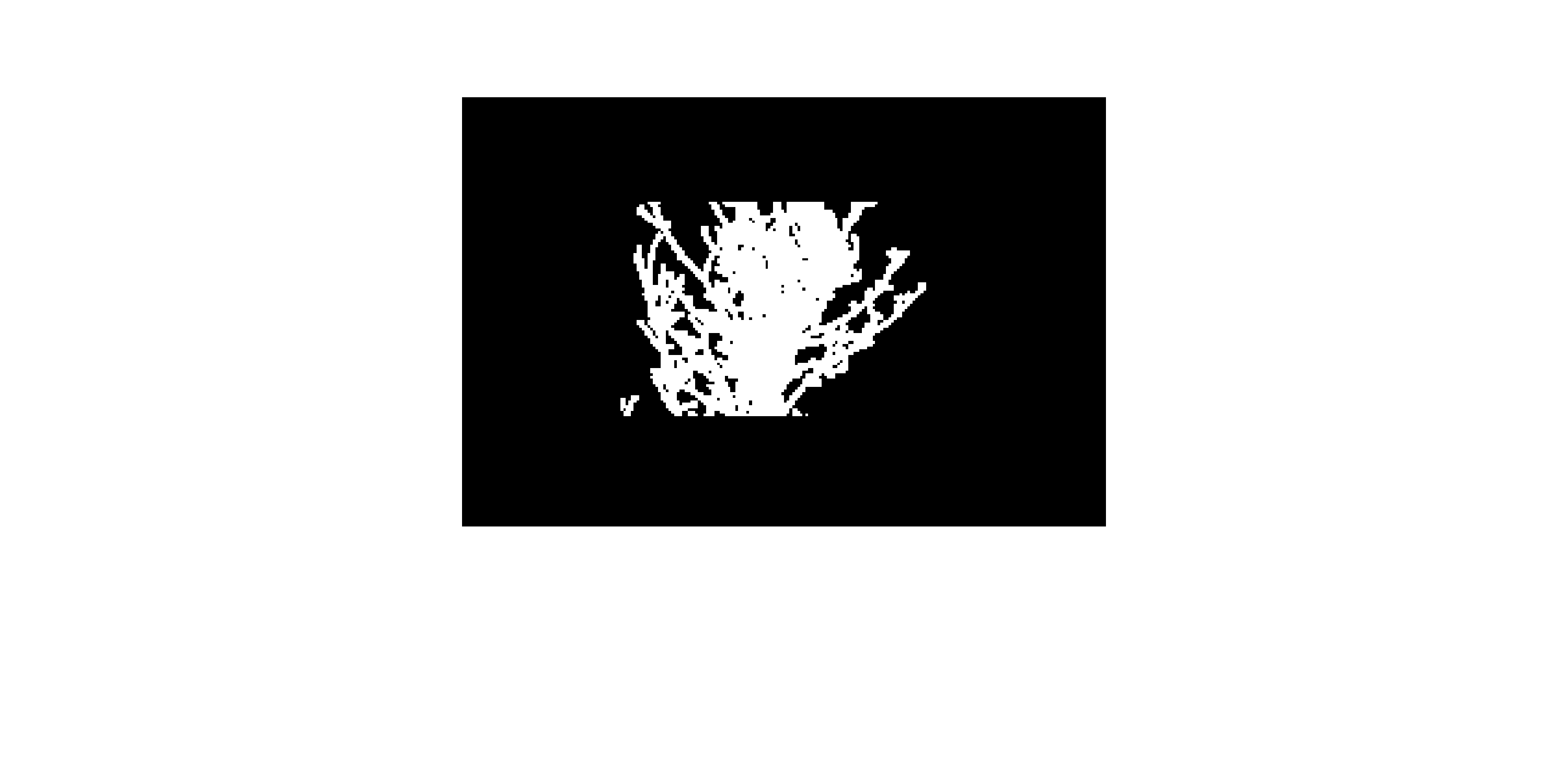, width=0.34\textwidth, bb= 430 230 1000 700, clip=}}&
\subfigure[]{\epsfig{file=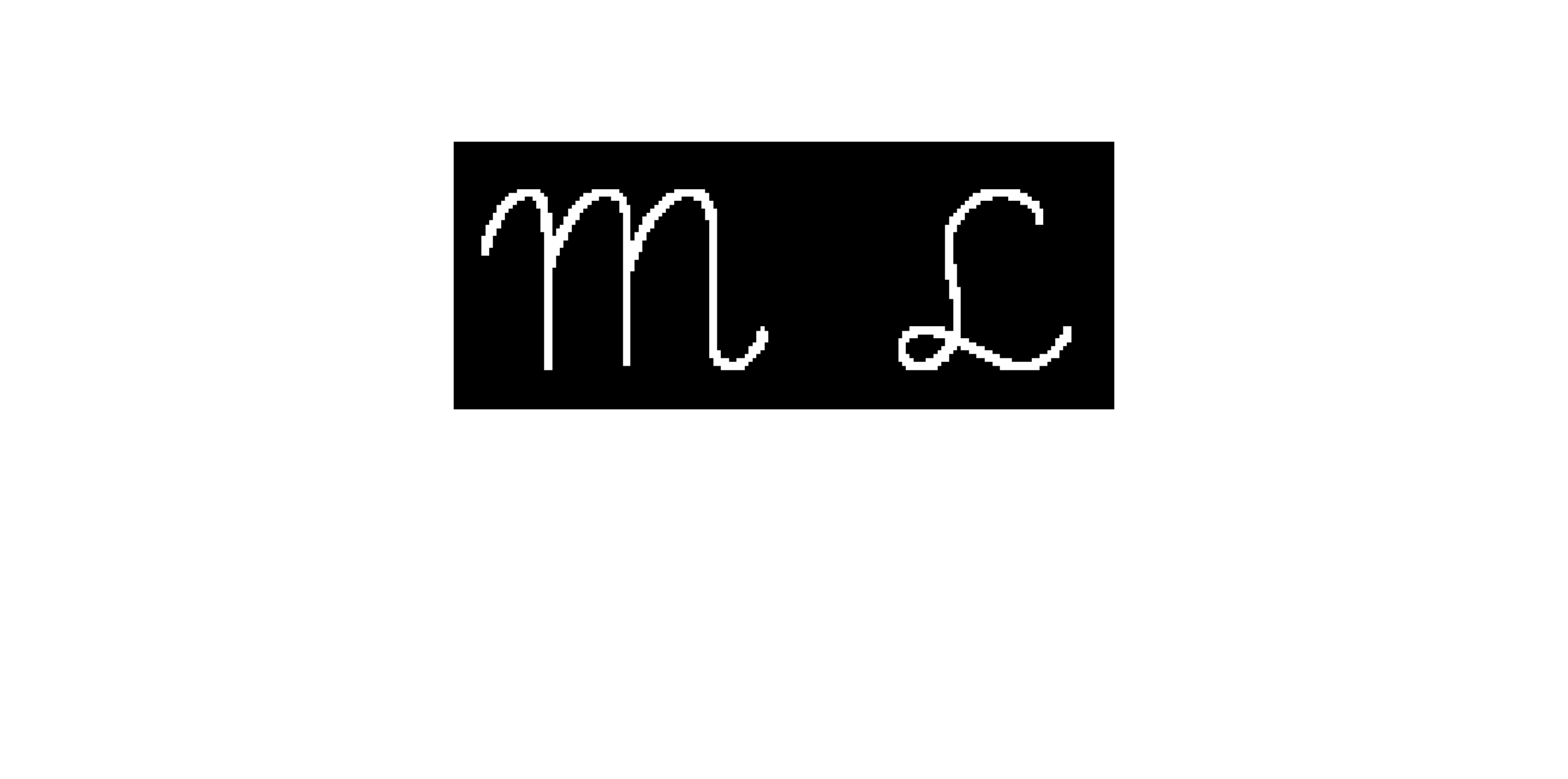, width=0.34\textwidth, bb= 400 335 1100 700, clip=}}\\
\end{tabular}
\caption{Images used for experiments, (a) phantom, (b) transaxial CT, (c) trees, (d) letters.}
\label{fig:img}
\end{figure}

\subsection{Sparse Signal Recovery} 

We first show sparse signal recovery performance when the signal has a sparse representation. The first experiment is on image data, where the non-zero elements constitute the content of an image, and exhibit continuity in the true signal $\boldsymbol{x}$. The images we used include the phantom, a CT scan, an image for trees, an image for letters (see Figure \ref{fig:img}), and MNIST dataset handwritten digits. The second experiment is on signals defined on graph structures, where the non-zero elements are distributed among the nodes of a graph. We used synthetic data defined on a binary tree, and data that are collected from IEEE 118-bus power system and IEEE 1354-bus power system, where the true signal $\boldsymbol{x}$ consists of the values of the state variables of a power system. Since the algorithms do not depend on the signal structure to find the non-zeros, they worked well with both types of data. Table \ref{tbl:img} and Table \ref{tbl:graph} show the performance of the proposed OMP-SR and BSR, and we report the number of iterations, the recovered optimal atoms, normalized MSE (NMSE), and running time in seconds. Image data are reported in Table \ref{tbl:img}, and graph data are reported in Table \ref{tbl:graph}.

If a $k$-sparse signal can be recovered by OMP within $k$ iterations, it can certainly be recovered by BSR within $k$ iterations. Those that cannot be recovered by OMP within $k$ iterations are shown to take far less than $k$ iterations and far less time by the BSR algorithm to fully recover. Since OMP-SR picks the same atoms as OMP does, we reported the result of OMP-SR in the same row as OMP and only reported its time (in blue text). Similarly, we report BSR  in the same row as gOMP, and report its running time (in blue text). It is observed that OMP-SR is faster than OMP, and BSR is faster than gOMP. The block size $c$ in BSR is a hyper parameter searched from \{2, 3, 4, 8\}.

\begin{figure}[!htbp]
\centering
\begin{tabular}{cc}
\subfigure[]{\epsfig{file=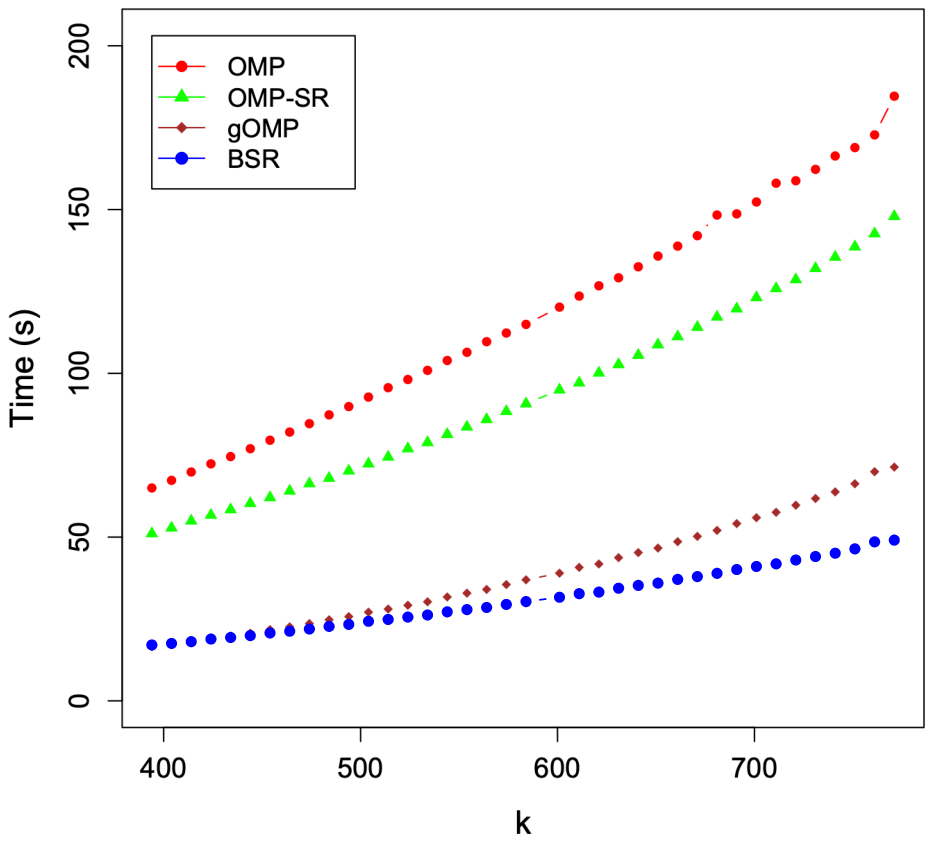, width=0.49\textwidth,bb= 50 240 600 650, clip=}}&
\subfigure[]{\epsfig{file=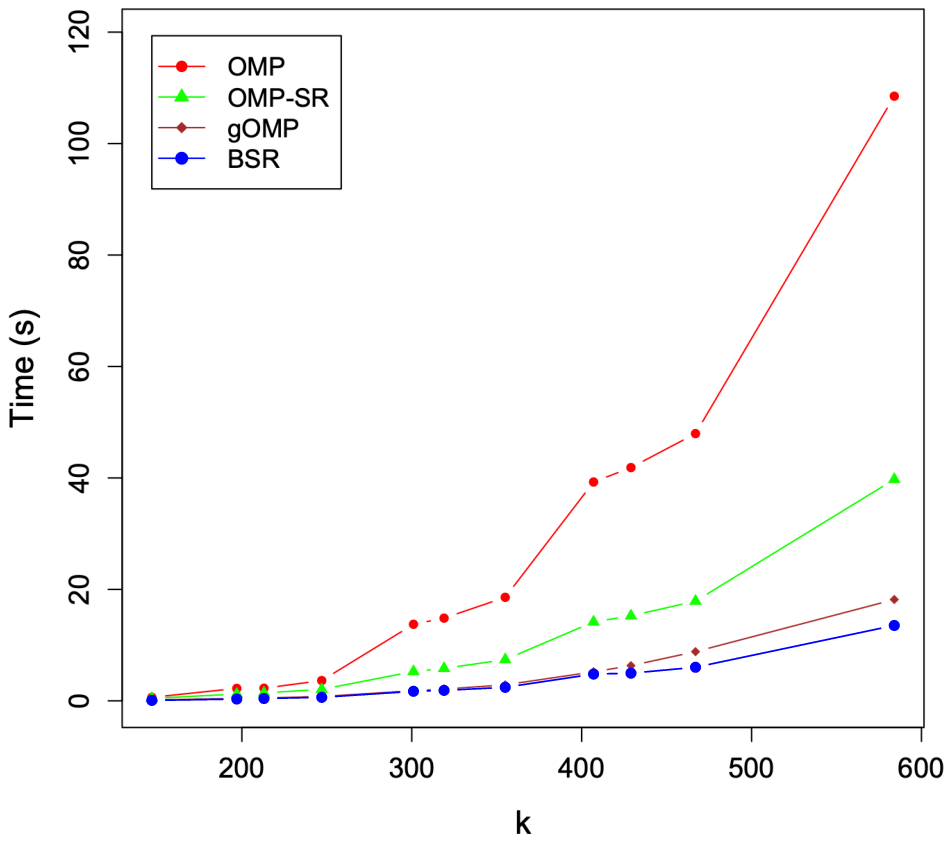, width=0.49\textwidth,bb= 50 240 600 650, clip=}}\\
\subfigure[]{\epsfig{file=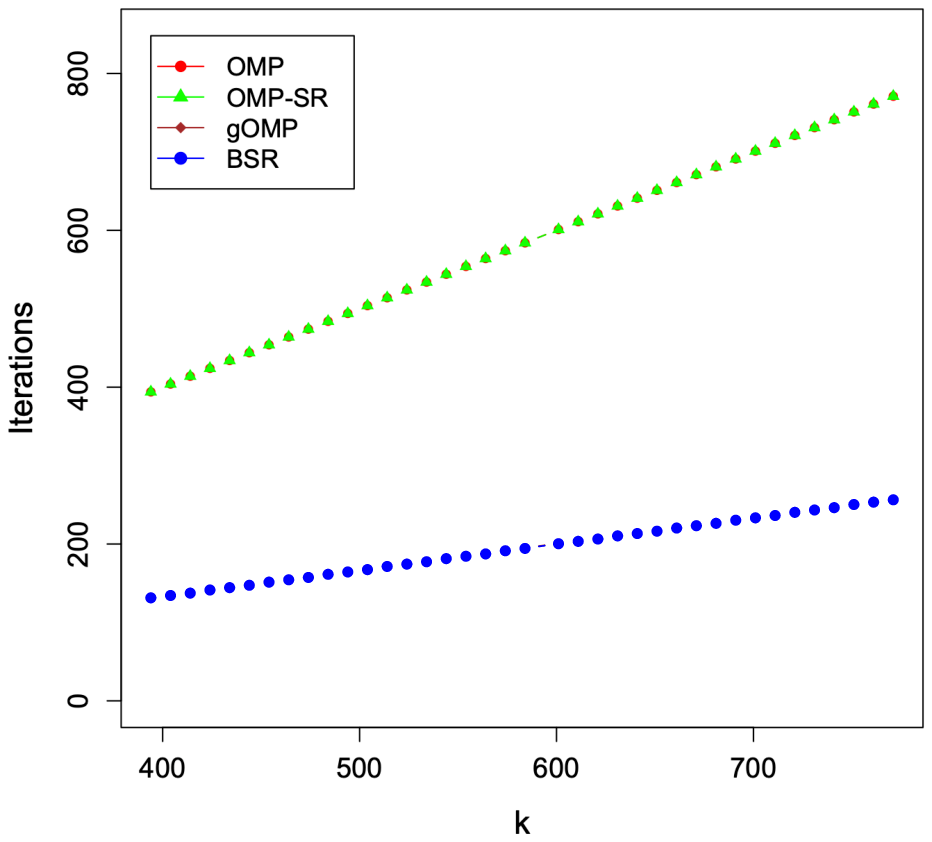,  width=0.49\textwidth,bb= 50 240 600 650, clip=}}&
\subfigure[]{\epsfig{file=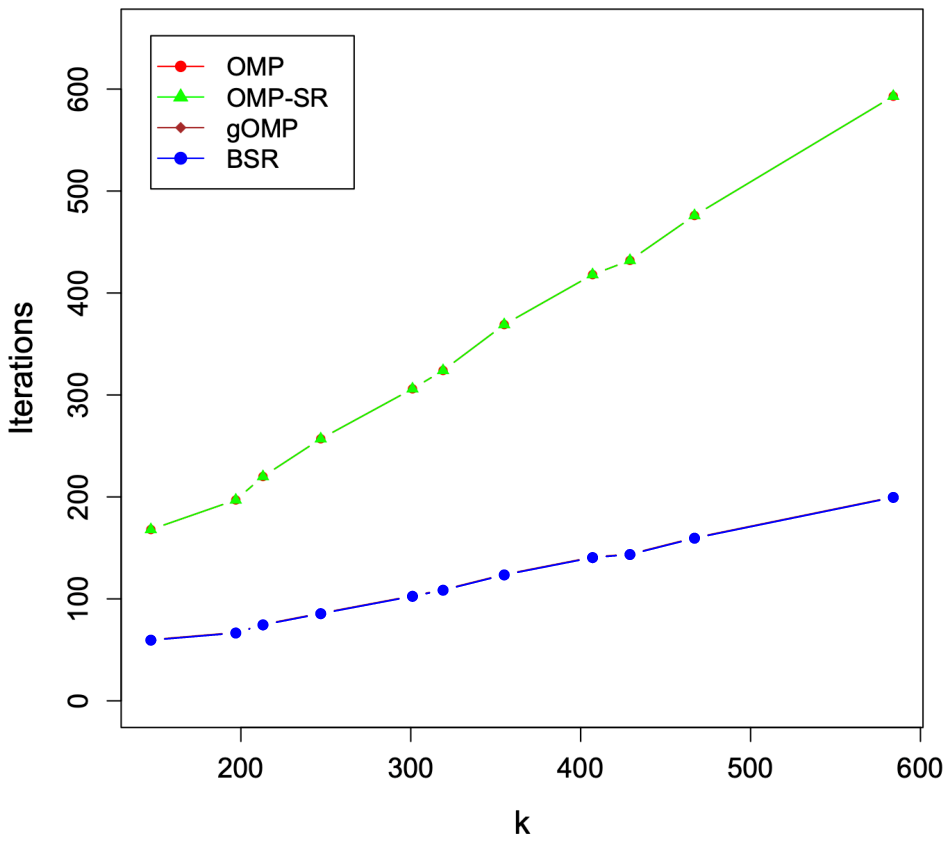, width=0.49\textwidth,bb= 50 240 600 650, clip=}}\\
\end{tabular}
\caption{Running time (top) and iterations (bottom) used by the algorithms to recover $k$ non-zeros in the signal. OMP and OMP-SR use the same number of iterations, and gOMP and BSR use the same number of iterations. The datasets used: (a),(c) phantom; (b),(d) MNIST dataset handwritten digit `7'.}
\label{fig:k}
\end{figure}

The third experiment is to show the relation between $k$ and running time. Image data for the phantom and the MNIST handwritten digit '7' were used. We created different versions from the original image to have different image sizes $d$ and different $k/d$ ratios. Figure \ref{fig:k} shows how the running time and iteration number increase as the number of non-zeros $k$ increases. The number of iterations is reduced by several folds in the blocked version, which is shown in (c) and (d). BSR is faster than gOMP per iteration, however, due to the reduced number of iterations, the advantage of BSR over gOMP becomes less significant compared to the advantage of OMP-SR over OMP, as the iteration number is reduced significantly.

\subsection{Sparse Approximation}
%\vspace{-0.05in}
The fourth experiment is for the sparse approximation of general signals with noises. We add noise $\bs \varepsilon $ to the Phantom image, and report  approximation errors when the measurements are subject to increasing levels of noise. Table \ref{tbl:approx} shows that at each noise level, BSR found the $k$ non-zeros with fewer iterations than OMP and significantly less running time.

\begin{table}[!ht]
\caption{Image datasets. Reported NMSE and time in seconds. Running time of our methods is highlighted in blue. OMP-SR is faster than OMP, and BSR is faster than gOMP. }
\label{tbl:img}
\scalebox{1.0}{
\begin{tabular}{|l|l|l|l|l|l|l|}
\toprule
Data & k & \!\!method \!\!\! & ite & \!\! found \!\!\! & NMSE & time \\
\midrule
\multirow{3}{*}{\begin{tabular}[c]{@{}l@{}}MNIST(3)\\$392 \!\! \times \!\! 784 \!\!$\end{tabular}} & \multirow{3}{*}{126} & OMP {\color{blue}  (OMP-SR)} & 126 & 110 & $0.0656$ & 0.4077 {\color{blue}(0.2977)} \\
\cmidrule{3-7}
&& OMP {\color{blue}  (OMP-SR)} & 142 & 126 & $<\!\!1e \!\!-\!\! 11$ & 0.5799 {\color{blue} (0.3280)}\\
\cmidrule{3-7}
&& gOMP {\color{blue} (BSR)} & 51 & 126 & $<\!\!1e \!\!-\!\! 11$& 0.2167 {\color{blue}(0.1298)}\\
\midrule
\multirow{3}{*}{\begin{tabular}[c]{@{}l@{}}MNIST(5)\\$392 \!\! \times \!\! 784\!\!$\end{tabular}} & \multirow{3}{*}{162} & OMP {\color{blue}  (OMP-SR)} & 162 & 99 & $0.7058$ & 0.5902 {\color{blue}(0.4204)} \\
\cmidrule{3-7}
&& OMP {\color{blue}  (OMP-SR)} & 784 & 162 & $<\!\!1e \!\!-\!\! 11$ & 18.0441 {\color{blue} (12.9938)}\\
\cmidrule{3-7}
&& gOMP {\color{blue} (BSR)} & 85 & 162 & $<\!\!1e \!\!-\!\! 11$& 0.5632 {\color{blue} (0.3926)}\\
\midrule
\multirow{3}{*}{\begin{tabular}[c]{@{}l@{}}MNIST(8)\\$392 \!\! \times \!\! 784\!\!$\end{tabular}} & \multirow{3}{*}{174} & OMP {\color{blue}  (OMP-SR)} & 174 & 92 & $0.8275$ & 0.7703 {\color{blue}(0.4934)} \\
\cmidrule{3-7}
&& OMP {\color{blue}  (OMP-SR)} & 546 & 174 & $<\!\!1e \!\!-\!\! 11$ & 11.0087 {\color{blue} (6.5816)}\\
\cmidrule{3-7}
&& gOMP {\color{blue} (BSR)} & 84 & 174 & $<\!\!1e \!\!-\!\! 11$& 0.5570 {\color{blue} (0.3855)}\\
\midrule
\multirow{3}{*}{\begin{tabular}[c]{@{}l@{}}MNIST(9)\\$ {392 \!\! \times \!\! 784}$\end{tabular}} & \multirow{3}{*}{130} 
& OMP {\color{blue}  (OMP-SR)} & 130 & 125 & 0.0727 & 0.4453 {\color{blue} (0.3190)} \!\! \\
\cmidrule{3-7}
&& OMP {\color{blue}  (OMP-SR)} & 135 & 130 & $<\!\!1e \!\!-\!\! 11$ & 0.5620 {\color{blue}(0.3917)} \\
\cmidrule{3-7}
&& gOMP {\color{blue} (BSR)}& 35 & 130 & $<\!\!1e \!\!-\!\! 11$ & 0.1521 {\color{blue} (0.1111)}\\
\midrule
\midrule

\multirow{3}{*}{\begin{tabular}[c]{@{}l@{}}Phantom\\${4512 \!\! \times \!\! 9024}$\end{tabular}} & \multirow{3}{*}{641} 
& OMP {\color{blue}  (OMP-SR)} & 641 & 638 & 0.0278 & 89.0610 {\color{blue} (70.8633)} \!\! \\
\cmidrule{3-7}
&& OMP {\color{blue}  (OMP-SR)} & 644 & 641 & $<\!\!1e \!\!-\!\! 11$ & 97.9924 {\color{blue}(75.9530)} \\
\cmidrule{3-7}
&& gOMP {\color{blue} (BSR)} & 81 & 641 & $<\!\!1e \!\!-\!\! 11$ & 13.0361 {\color{blue} (9.1253)}\\
\midrule
\multirow{3}{*}{\begin{tabular}[c]{@{}l@{}}\!\! Transaxial CT \!\! \!\! \\${4225 \!\! \times \!\! 8450}$\end{tabular}} & \multirow{3}{*}{\!\! 1089 \!\!} 
& OMP {\color{blue}  (OMP-SR)} & 1089 & 1064 & $ 0.0675$ & 282.6071 {\color{blue}(250.1557)} \!\! \\
\cmidrule{3-7}
&& OMP {\color{blue}  (OMP-SR)}  & 1115 & 1089 & $<\!\!1e \!\!-\!\! 11$ & 302.0011 {\color{blue}(263.0615)}\\
\cmidrule{3-7}
&& gOMP {\color{blue} (BSR)} & 57 & 1089 & $<\!\!1e \!\!-\!\! 11$ & 17.5773 {\color{blue} (14.0619)}\\
\midrule
\multirow{3}{*}{\begin{tabular}[c]{@{}l@{}}{\color{black} Trees}\\${19200 \!\! \times \!\! 38400 \!\!}$\end{tabular}} & \multirow{3}{*}{\!\! 4670 \!\!}
& OMP {\color{blue}  (OMP-SR)} & 4670 & 4652 & $\!\! 0.00114$ \!\!\! & \!\! 2391.0661 \!\! {\color{blue}(702.4279)} \!\!\!\!\\
\cmidrule{3-7}
&& OMP {\color{blue}  (OMP-SR)} & 4688 & 4670 & $<\!\!1e \!\!-\!\! 11$ & \!\! 2571.1333 \!\! {\color{blue}(754.8234)} \!\!\!\! \\
\cmidrule{3-7}
&& gOMP {\color{blue} (BSR)} & 117 & 4670 & $<\!\!1e \!\!-\!\! 11$ & 23.0564 {\color{blue} (20.0294)} \\
\midrule
\multirow{3}{*}{\begin{tabular}[c]{@{}l@{}}Letters\\${5712 \!\! \times \!\! 11424}$\end{tabular}} & \multirow{3}{*}{851} 
& OMP {\color{blue}  (OMP-SR)} & 851 & 851 & $<\!\!1e \!\!-\!\! 11$ & 191.7733 {\color{blue}(129.9197)} \!\!\\
\cmidrule{3-7}
&& gOMP {\color{blue} (BSR)} & 107 & 851 & $<\!\!1e \!\!-\!\! 11$ & 20.0811 {\color{blue} (16.4433)}\\
\bottomrule
\end{tabular}
}
\end{table}

\begin{table}[!htbp]
\caption{Synthetic data for signals defined on graph structures. Reported NMSE and time in seconds. Running time of our methods is highlighted in blue. OMP-SR is faster than OMP, and BSR is faster than gOMP.}
\label{tbl:graph}
\scalebox{1.0}{
\begin{tabular}{|l|l|l|l|l|l|l|}
\toprule
Data & k & method & ite & found & NMSE & time \\
\midrule
\multirow{3}{*}{\begin{tabular}[c]{@{}l@{}}Binary Tree\\${256 \!\! \times \!\! 512}$\end{tabular}} & \multirow{3}{*}{70 } & 
OMP {\color{blue}  (OMP-SR)} & 70 & 70 & $<\!\!1e \!\!-\!\! 11$ &  0.1105 {\color{blue}(0.0798)} \!\! \\ 
\cmidrule{3-7}
&& gOMP {\color{blue} (BSR)} & 25 & 70 & $<\!\!1e \!\!-\!\! 11$ & 0.02778 {\color{blue}(0.024294)} \\
\midrule

\multirow{3}{*}{\begin{tabular}[c]{@{}l@{}}118 Bus\\${59 \!\! \times \!\! 118}$\end{tabular}} & \multirow{3}{*}{14} 
& OMP {\color{blue}  (OMP-SR)} & 14 & 14 & $<\!\!1e \!\!-\!\! 11$ &  0.0284 {\color{blue}(0.0172)} \!\! \\
\cmidrule{3-7}
&& gOMP {\color{blue} (BSR)} & 5 & 14 & $<\!\!1e \!\!-\!\! 11$ & 0.00609 {\color{blue} (0.003876)}\\
\midrule
\multirow{3}{*}{\begin{tabular}[c]{@{}l@{}}118 Bus\\${59 \!\! \times \!\! 118}$\end{tabular}} & \multirow{3}{*}{100} 
& OMP {\color{blue}  (OMP-SR)} & 100  & 86  & 0.8406  & 0.0706 {\color{blue}(0.0582)} \!\! \\
\cmidrule{3-7}
&& OMP {\color{blue}  (OMP-SR)} & {118}  & 100  & $<\!\!1e \!\!-\!\! 11$  & 0.0922 {\color{blue}(0.0659)}\\
\cmidrule{3-7}
&& gOMP {\color{blue} (BSR)} &  40 & 100  & $<\!\!1e \!\!-\!\! 11$  & 0.03400 {\color{blue} (0.025842)} \\
\midrule

\multirow{3}{*}{\begin{tabular}[c]{@{}l@{}}1354 Bus\\${677 \!\! \times \!\! 1354}$\end{tabular}} & \multirow{4}{*}{270} 
& OMP {\color{blue}  (OMP-SR)} & 270 & 215 & $0.2029$ & 3.7723 {\color{blue}(2.1686)} \!\! \\
\cmidrule{3-7}
&& OMP {\color{blue}  (OMP-SR)} & 336 & 270 & $<\!\!1e \!\!-\!\! 11$ & 5.4101 {\color{blue}(3.4590)}\\
\cmidrule{3-7}
&& gOMP {\color{blue} (BSR)} & 96 & 270 & $<\!\!1e \!\!-\!\! 11$ & 1.2641 {\color{blue} (0.982119)}\\
\bottomrule
\end{tabular}
}
\end{table}

\begin{table}[!htbp]
\caption{Results for the phantom image with increasing noise level $\| \bs \varepsilon\|_2$. Reported normalized approximation error $\frac{\| \bs y - \Phi \bs x \|_2}{\|\bs y\|_2}$, and running  time in seconds. Running time of our methods is highlighted in blue. OMP-SR is faster than OMP, and BSR is faster than gOMP.}
\label{tbl:approx}
\scalebox{1.0}{
\begin{tabular}{|l|l|l|l|l|l|l|}
\toprule
Noise & k &\!\! method \!\!& ite & \!\! found &\!\! \!\! $\frac{\| \bs y - \Phi \bs x \|_2}{\|\bs y\|_2}$ \!\! & time \\
\toprule
\multirow{3}{*}{\begin{tabular}[c]{@{}l@{}} $\| \bs \varepsilon\|_{2}=0.1$\end{tabular}} & \multirow{3}{*}{641} & OMP {\color{blue}  (OMP-SR)} & 641 & 638 & 0.0231 & 88.7496 {\color{blue} (70.3730)}\\
\cmidrule{3-7}
&& OMP {\color{blue}  (OMP-SR)} & 644 & 641 & $0.0001$ & 94.6122 {\color{blue}(72.1749)} \\
\cmidrule{3-7}
&& gOMP {\color{blue} (BSR)}  & 81 & 641 & $0.0001$ & 11.3051 {\color{blue} (9.1572)}\\
\midrule
\multirow{3}{*}{\begin{tabular}[c]{@{}l@{}}$\| \bs \varepsilon\|_{2}\! = \! 50$\end{tabular}} & \multirow{3}{*}{641} & OMP {\color{blue}  (OMP-SR)} & 641 & 638 & 0.0777 & 87.1276 {\color{blue} (70.6804)}\\
\cmidrule{3-7}
&& OMP {\color{blue}  (OMP-SR)} & 644 & 641 & $0.0727$ & 95.4393 {\color{blue}(71.5545)} \\
\cmidrule{3-7}
&& gOMP {\color{blue} (BSR)} & 81 & 641 & $0.0720$ & 11.7429 {\color{blue} (9.7602)}\\
\midrule
\multirow{3}{*}{\begin{tabular}[c]{@{}l@{}} $\| \bs \varepsilon\|_{2}\!=\!100$\end{tabular}} & \multirow{3}{*}{641} & OMP {\color{blue}  (OMP-SR)} & 641 & 631 & 0.1470 & 88.9356 {\color{blue} (72.0881)}\\
\cmidrule{3-7}
&& OMP {\color{blue}  (OMP-SR)} & 684 & 641 & $0.1337$ & 165.5339 {\color{blue}(138.7913)} \\
\cmidrule{3-7}
&& gOMP {\color{blue} (BSR)}  & 101 & 641 & $0.1287$ & 16.0534 {\color{blue} (13.8060)}\\
\midrule
\multirow{3}{*}{\begin{tabular}[c]{@{}l@{}} $\| \bs \varepsilon\|_{2}\!=\! 150$\end{tabular}} & \multirow{3}{*}{641} & OMP {\color{blue}  (OMP-SR)} & 641 & 621 & 0.2687 & 89.2221 {\color{blue} (70.3730)}\\
\cmidrule{3-7}
&& OMP {\color{blue}  (OMP-SR)} & 789 & 641 & $0.1671$ &  134.5601  {\color{blue}(101.7830)} \\
\cmidrule{3-7}
&& gOMP {\color{blue} (BSR)} & 197 & 641 & $0.1671$& 27.9475 {\color{blue} (25.4323)}\\
\midrule
\end{tabular}
}
\end{table}

\section{Discussion and Future Work}
\label{sec:conclusion}

OMP has the advantage of simplicity. A greedy algorithm such as OMP is easy to implement but difficult to analyze. This work offered significant performance improvement over the classical OMP and its extension gOMP with theoretical analysis for convergence and approximation error bound. In addition, the proposed changes for OMP come from a principled approach. They work well when combined with other heuristic or ensemble approaches. One possible future work direction is to improve the greedy choice by leveraging the structure in the signal model. 

In addition, the minimal $\ell_1$ norm solution is the sparsest only when the signal is sparse enough \cite{BP-sparse}. Therefore, another future work direction is to identify the specific measurement matrix property that drives sparsity during $\ell_1$ norm minimization and use that to improve the greedy choice in an iterative procedure.

\bibliographystyle{splncs04}
\bibliography{CS,added}

\appendix
\section*{\centering{Appendix}}

\section{Proof for Theorem \ref{thm:strong}}

\setcounter{theorem}{0}
\begin{theorem}[The strong exact recovery condition for BSR] 
A sufficient condition for BSR to recover a $k$-sparse signal within $\lceil k/c\rceil $ iterations is that
\begin{equation} \nonumber
\rho_c(\boldsymbol{r}) < 1
\end{equation}
holds for all iterations.
\end{theorem}

{\bf Proof}: 
In all but the last iteration, BSR algorithm will select $c$ columns from $\Phi_\mathrm{opt}$ that have not been selected by the algorithm. This can be observed from two aspects: 1) the columns that have been selected so far will not contribute to the maximum $\ell_2$-norm  $\max\limits_{\Omega_2}\left\| \Phi^\top_{\Omega_2}\boldsymbol{r}\right\|_{2}$ since the residual $\boldsymbol{r}$ is orthogonal to them, therefore the maximum value have to be achieved by the new columns the algorithm has not seen so far; and  2) with $\rho_c(\boldsymbol{r})<1$, the BSR algorithm will select as many optimal columns as possible in each iteration. Therefore the $c$ columns selected by the algorithm in each iteration (except the last) will not include any column from $\Psi$, and they all  have to come from $\Phi_\mathrm{opt}$. In the last iteration it is possible to select non-optimal columns from $\Psi$ because all optimal columns in $\Phi_\mathrm{opt}$ have been exhausted. Therefore the BSR algorithm is guaranteed to locate $k$ optimal columns in $\lceil k/c\rceil$ iterations.
\qed

\section{Proof for Theorem \ref{thm:weak}}

\setcounter{theorem}{1}

\begin{theorem}[The Weak Exact Recovery Condition for BSR]
A sufficient condition for BSR to recover a $k$-sparse signal within $k$ iterations is that
\begin{equation} \nonumber
\rho(\boldsymbol{r}) < 1
\end{equation}
holds for all iterations.
\end{theorem}

\vspace{0.1in}

{\bf Proof}:
If $\rho(\boldsymbol{r}) <1$ holds, then in each iteration BSR can choose at least one optimal column from $\Phi_\text{opt}$. This is because the BSR algorithm chooses the $c$ columns that give the maximum $\ell_2$-norm  $\max\limits_{|\Omega|=c}\left\|{{(\Phi_{\mathrm{opt}}})^\top_\Omega}\boldsymbol{r}\right\|_{2}$. The $c$ columns that achieved the maximum $\ell_2$-norm have the largest $c$ absolute inner products with residual $\boldsymbol{r}$, which certainly include the largest value. If the condition $\rho(\boldsymbol{r}) < 1$ holds in all iterations, the BSR algorithm is guaranteed to select at least one new column from $\Phi_{\mathrm{opt}}$. Therefore, for a $k$-sparse signal, BSR can find all optimal columns in at most $k$ iterations. \qed
 
\section{Proof for Lemma \ref{eq:lm-rho}}

\setcounter{lemma}{0}

\begin{lemma} 
If  $\max\limits_{\boldsymbol{\psi}} \left\| \left(X^{+} \right)_{\Pi,:} \boldsymbol{\psi}\right\|_1 <1$, where vector $\boldsymbol{\psi}$ ranges over columns of $\Psi_{\mybar{J}}$, then the residual $\boldsymbol{r}$ satisfies $\rho(\boldsymbol{r})<1$.
\end{lemma}

{\bf Proof}: 

$\begin{aligned} 
\rho(\boldsymbol{r}) 
&=\frac{\left\|\Psi^{\top} \boldsymbol{r}\right\|_{\infty}}{\left\|\Phi_{\mathrm{opt}}^{\top} \boldsymbol{r}\right\|_{\infty}} \\ 
&\overset{(a)}=\frac{\left\|\Psi_{\mybar{J}}^{\top} \boldsymbol{r}\right\|_{\infty}}{\left\|\Phi_{\mathrm{opt}}^{\top} \boldsymbol{r}\right\|_{\infty}} \\ 
& \overset{(b)}=\frac{\left\|\Psi_{\mybar{J}}^{\top}   X(X^\top X)^{-1} X^\top \boldsymbol{r}\right\|_{\infty}} {\left\|\Phi_{\mathrm{opt}}^{\top} \boldsymbol{r}\right\|_{\infty}} \\ 
& \overset{(c)}=\frac{\left\|\Psi_{\mybar{J}}^{\top} \left( X(X^\top X)^{-1}  \right)_\Pi X_\Pi^{\top} \boldsymbol{r}\right\|_{\infty}} {\left\|\Phi_{\Pi}^{\top} \boldsymbol{r}\right\|_{\infty}} \\ 
& \overset{(d)} \leq\left\|\Psi_{\mybar{J}}^{\top} \left(\left(X^{+}\right)^{\top} \right)_\Pi\right\|_{\infty} \\
& \overset{(e)}=\left\| \left(X^{+}\right)_{\Pi,:} \Psi_{\mybar{J}} \right\|_{1} \\ 
&\overset{(f)}=\max _{\boldsymbol{\psi}}\left\| \left(X^{+}\right)_{\Pi,:} \boldsymbol{\psi}\right\|_{1} ,\\
\end{aligned}$

where $\boldsymbol{\psi}$ ranges over columns of $\Psi_{\mybar{J}}$.

$(a)$ is because $\left\|\Psi^{\top} \boldsymbol{r}\right\|_{\infty} =\left\|\Psi_{\mybar{J}}^{\top} \boldsymbol{r}\right\|_{\infty}$, since columns in $\Psi_J$ are orthogonal to $\boldsymbol{r}$.

$(b)$ is due to that residual $\boldsymbol{r}$ lies in the column span of $X$. Sparse representation of $\boldsymbol{y}= \sum\limits_{j \in \Lambda_\mathrm{opt} } a_j  \bs \varphi_j $, so $\boldsymbol{y}$ lies in the column span of $\Phi_\mathrm{opt}$. Since $X$ includes all columns of $\Phi_\mathrm{opt}$, $\boldsymbol{y}$ also lies in the column span of $X$. On the other hand, $X$  includes all the columns found by the algorithm so far, so $\Phi_\Gamma \boldsymbol{s}_\Gamma$ also lies in the column span of $X$. Therefore the residual $ \boldsymbol{r}=\boldsymbol{y}-\Phi_\Gamma \boldsymbol{s}_\Gamma$ also lies in the column span of $X$.

$X(X^\top X)^{-1}X^\top$ is a projector onto the column span of $X$. Since $\boldsymbol{r}$ lies in the column span of $X$, so the projection of $\boldsymbol{r}$ onto the column span of $X$ is $\boldsymbol{r}$ itself: \[X(X^\top X)^{-1}X^\top \boldsymbol{r} = \boldsymbol{r}. \]

 In the vectors $X^\top \boldsymbol{r}$ and $\Phi_\mathrm{opt}^\top \boldsymbol{r}$, all the nonzero elements are from $\Phi^\top_\Pi \boldsymbol{r}$. Thus,
\[ \left\| X^\top \boldsymbol{r} \right\|_\infty=\left\|\Phi^\top_\mathrm{opt} \boldsymbol{r} \right\|_\infty= \left\|X^\top_\Pi \boldsymbol{r} \right\|_\infty= \left\|\Phi^\top_\Pi \boldsymbol{r} \right\|_\infty . \]

This leads to $(c)$.
 
$(d)$ is direct from the definition of matrix norm. Suppose that $A$ is a matrix, and $ \boldsymbol{v} $ is a vector with nonzero components. For $1 \le p \le \infty$, we have

\[ \|A\|_p=\sup\limits_{\boldsymbol{v} \ne 0} \frac{\|A \boldsymbol{v} \|_p}{\| \boldsymbol{v} \|_p} .\]

Therefore, 
\begin{equation}
\label{eq:p-norm}
 \|A \boldsymbol{v} \|_p \le  \|A\|_p  \| \boldsymbol{v} \|_p. 
\end{equation}

$(e)$ is due to that the matrix norm $\| \cdot \|_{\infty}$ takes the maximum absolute row sum, and $\| \cdot \|_{1}$ takes the maximum absolute column sum, so $\| A^\top \|_{\infty}=\| A \|_{1}$ for a given matrix $A$. 

$(f)$ follows the definition of matrix norm $\| \cdot \|_{1}$ to get the maximum absolute column sum of the matrix.

Therefore, if $\max\limits_{\boldsymbol{\psi}}\left\| \left(X^{+}\right)_{\Pi,:} \boldsymbol{\psi}\right\|_{1}<1$, where $\boldsymbol{\psi}$ ranges over columns of $\Psi_{\mybar{J}}$, then $\rho(\boldsymbol{r})<1$. \qed

\section{Proof for Lemma \ref{eq:lm-mu}}

\begin{lemma}
$\max\limits_{\boldsymbol{\psi}\in \Psi_{\mybar{J}}} \left\| \left(X^{+}\right)_{\Pi,:} \boldsymbol{\psi}\right\|_1 <1$ whenever $\mu_1(l)+\mu_1(n)<1$ holds, where $n$ is the number of columns in $X$, and $l =\min ( | \Pi |, k-1 )$.
\end{lemma}

{\bf Proof:}

$\begin{aligned}
     &\max _{\boldsymbol{\psi} \in \Psi_{\mybar{J}}}\left\| \left(X^{+}\right)_{\Pi,:} \boldsymbol{\psi}\right\|_{1}\\
=   &\max _{\boldsymbol{\psi} \in \Psi_{\mybar{J}}}\left\| \left( \left(X^{\top} X\right)^{-1} X^{\top} \right)_{\Pi,:}\boldsymbol{\psi}\right\|_{1}\\
=   &\max _{\boldsymbol{\psi} \in \Psi_{\mybar{J}}}\left\| \left( \left(X^{\top} X\right)^{-1} \right)_{\Pi,:} X^{\top} \boldsymbol{\psi}\right\|_{1}\\
\le & \left\| \left( \left(X^{\top}X\right)^{-1} \right)_{\Pi,:} \right\|_{1} \cdot \left(\max _{\boldsymbol{\psi} \in \Psi_{\mybar{J}}}\left\|X^{\top}\boldsymbol{\psi}\right\|_{1}\right)\\
\le  & \frac{\mu_1(n)}{1-\mu_1(l)}, \\
\end{aligned}$

\vspace{5pt}

where $n$ is the number of columns in $X$, and $l =\min ( | \Pi |, k-1 )$. 

\vspace{5pt}

It can be shown that
\begin{enumerate}
\item[1).] $\max\limits_{\boldsymbol{\psi} \in \Psi_{\mybar{J}}}\left\|X^{\top}\boldsymbol{\psi}\right\|_{1}  \le \mu_1(n)$, and
\item[2).] $\left\| \left( \left(X^{\top}X\right)^{-1} \right)_{\Pi,:} \right\|_{1}  \le \frac{1}{1-\mu_1(l)}$.
\end{enumerate}

1) can be proved following the definition of $\mu_1(\cdot)$,

\[\begin{aligned}
\max _{\boldsymbol{\psi} \in \Psi_{\mybar{J}}}\left\|X^{\top}\boldsymbol{\psi}\right\|_{1} 
& = \max_{\boldsymbol{\psi} \in \Psi_{\mybar{J}}} \sum_{\boldsymbol{\varphi_j} \in X} |\langle  \boldsymbol{\varphi}_j, \boldsymbol{\psi} \rangle| \\
& \le \mu_1(n). \\
\end{aligned}\]

To prove 2), recall that the columns in the dictionary $\Phi$ are normalized unit vectors (i.e., $\| \boldsymbol{\varphi} \|_2=1$, for $\boldsymbol{\varphi} \in \Phi$). We can represent \[X^\top X=I_n+A , \]
where $I_n$ is an identity matrix of size $n \times n$, and $A$ is a matrix of size $n \times n$ collecting the pairwise inner products, so $ A_{jk}=\langle \boldsymbol{\varphi}_j, \boldsymbol{\varphi}_k \rangle, \textrm{ for } j \ne k$, and the diagonal $A_{jj}=0$, and hence $\|A\|_1 \le \mu_1(n-1)$ following the definition of the cumulative coherence function $\mu_1(\cdot)$.

When $\|A\|_1 < 1$, the Neumann series $\sum\limits_{k=0}^\infty (-A)^k=(I_n+A)^{-1}$ (add reference here).
Therefore, $(X^\top X)^{-1}=(I_n+A)^{-1}=\sum\limits_{k=0}^\infty (-A)^k $. Note that having $\mu_1(n-1) <1 $ is sufficient to make $\|A\|_1 < 1$.

\[\begin{aligned}
\left\| \left( \left(X^{\top}X \right)^{-1} \right)_{\Pi,:}\right\|_{1} 
& = \left\|\left(\sum\limits_{k=0}^\infty (-A)^k\right)_{\Pi,:} \right\|_1 \\
&\overset{(a)} \le \sum\limits_{k=0}^\infty \|A_{\Pi,:}\|^k_{1}  \\
&\overset{(b)} = \frac{1}{1-\|A_{\Pi,:}\|_1} \\
&\overset{(c)} \le \frac{1}{1-\mu_1(l)} .\\
\end{aligned}\]

$(a)$ is due to $\|(-A)^k\|_1 \le \|A\|_1^k$;  $(b)$ is due to $\|A_{\Pi,:}\|_1 \le \|A\|_1 <1$; $(c)$ is due to  $\|A_{\Pi,:}\|_1 \le \mu_1(l)$. At the beginning before the first iteration, $ X=\Phi_\mathrm{opt}$, $\Pi=\Lambda_\mathrm{opt}$, and $A$ is a $k \times k$ matrix with zeros at the diagonal positions, so $\|A_{\Pi,:}\|_1 \le \mu_1(k-1)$. After the first iteration, $| \Pi | \le k-1$. Therefore,  $\|A_{\Pi,:}\|_1 \le \mu_1(l)$ for all iterations, with $l=\min(|\Pi|, k-1)$.

Therefore $\max\limits_{\boldsymbol{\psi}\in \Psi_{\mybar{J}}}\left\| \left(X^{+}\right)_{\Pi,:} \boldsymbol{\psi}\right\|_{1}<1$ whenever $\mu_1(l)+ \mu_1(n)<1$. \qed

\section{Proof for Theorem \ref{eq:thm-strong-mu}}

\begin{theorem}[The strong exact recovery condition for BSR]
Suppose that $\mu$ is the coherence of the dictionary as defined in \eqref{coherence}. A sufficient condition for BSR to recover a $k$-sparse signal within $\lceil k/c \rceil$ iterations is that
\begin{equation} \nonumber
\mu(2k-1)<1.
\end{equation}
\end{theorem}

To prove Theorem \ref{eq:thm-strong-mu}, we first show that $\mu_1(l)+ \mu_1(n)<1$ is also sufficient for $\rho_c({\bs r}) < 1 $ to hold in Theorem \ref{thm:strong}. 

\begin{lemma}
\label{lm-strong-mu}
$\rho_c({\bs r}) < 1$ whenever $\mu_1(l)+ \mu_1(n)<1$. 
\end{lemma}

{\bf Proof:}
From the definition of  $\rho_c({\bs r})$, it is suggested that $\Omega_1$ and $\Omega_2$ both can have columns from $\Psi$ and $\Phi_\mathrm{opt}$. If there is a column $\bs \varphi_j$ from  $\Phi_{\Omega_1}$ and a column $\bs \varphi_l$ from $\Phi_{\Omega_2}$ such that $|\langle \bs \varphi_j, \bs r \rangle| > |\langle \bs \varphi_l, \bs r \rangle| $, then the greedy algorithm would have selected $\bs \varphi_j$. In this case, we would switch the memberships of two columns in $\Omega_1$ and $\Omega_2$ as long as such switching does not violate the condition $|\Omega_2 \cap \Lambda_\mathrm{opt} | > |\Omega_1 \cap \Lambda_\mathrm{opt} | $. The only case that the switching is forbidden is that $\bs \varphi_j$ is a column in  $\Psi$ and $\bs \varphi_l$ is a column in $\Phi_\mathrm{opt}$ so that switching would have violated the above condition. There will be at least one such column in $\Psi$ and one column in $\Phi_\mathrm{opt} $ that cannot be switched.

Let $\Theta_1 \subseteq \Omega_1$ and $\Theta_2 \subseteq \Omega_2$ be the indices for such non-switchable columns in $\Psi$ and $\Phi_\mathrm{opt}$, respectively. We have $|\Theta_1|=|\Theta_2| = |\Omega_2 \cap \Lambda_\mathrm{opt} | - |\Omega_1 \cap \Lambda_\mathrm{opt} | \ge 1$. We define 
\begin{equation}
\rho'_c(\bs r) \stackrel{\text { def }}{\longeq}  \frac{\max\limits_{\Theta_1}\left\| \Psi^\top_{\Theta_1} {\bs r}\right\|_{2}}{\max\limits_{\Theta_2}\left\|  (\Phi_\mathrm{opt})^\top_{\Theta_2}\boldsymbol{r}\right\|_{2}},
\end{equation}

Note that $\rho'_c(\bs r) <1$ is sufficient for $\rho_c(\bs r) <1$, since for the other columns in $\Phi_{\Omega_1 \setminus \Theta_1}$, if it happens to have a larger absolute inner product with $\bs r$, we can switch it with a column in $\Phi_{\Omega_2}$ to improve the maximum value obtained by $\Omega_2$ in \eqref{eq:rho-c-def} since switching does not violate the condition $ |\Omega_2 \cap \Lambda_\mathrm{opt} | > |\Omega_1 \cap \Lambda_\mathrm{opt} | $.

Next we show that the sufficient condition for $\rho(\bs r) <1 $ is also sufficient for $\rho'_c(\bs r) <1 $.

 \begin{equation}
 \label{eq:rho-c-sum}
% \begin{aligned}
 \frac{\max\limits_{\Theta_1}\left\| \Psi^\top_{\Theta_1} {\bs r}\right\|_{2}}{\max\limits_{\Theta_2}\left\|  (\Phi_\mathrm{opt})^\top_{\Theta_2}\boldsymbol{r}\right\|_{2}}<1  \;\; \Longleftrightarrow  \;\;
 \frac{\max\limits_{\Theta_1}\sum\limits_{{\bs \psi}_j \in \Psi_{\Theta_1}}\left\| {\bs \psi}_j^\top{\bs r}\right\|_{2}^2}
{\max\limits_{\Theta_2}\sum\limits_{{\bs \varphi}_l \in (\Phi_\mathrm{opt})_{\Theta_2}}\left\| {\bs \varphi}_l^\top \boldsymbol{r}\right\|_{2}^2}<1. 
% \end{aligned}
\end{equation}

Now we sort $\Theta_1$ and $\Theta_2$ in the descending order of the absolute inner products $| \langle \bs \psi_j, \bs r \rangle |$ and $| \langle \bs \varphi_l, \bs r \rangle |$, respectively. Consequently,  $\Theta_1=\{ j_{(1)}, j_{(2)}, j_{(3)}, \ldots \}$, and $\Theta_2=\{ l_{(1)}, l_{(2)}, l_{(3)}, \ldots \}$ with subscript $(i)$ indicating the $i$-th largest value.

It is sufficient to show that $ \rho'_c(\bs r) <1 $ as long as each of the $ \rho'_{(i)}(\bs r) <1 $, for $i =1, \ldots, |\Theta_1|$, where $ \rho'_{(i)}(\bs r)$ is defined as

\begin{equation} 
\rho'_{(i)}(\bs r)\stackrel{\text { def }}{\longeq} \frac{|\langle {\bs \psi}_{ j_{(i)} }, {\bs r} \rangle |} { | \langle {\bs \varphi}_{ l_{(i)}}, {\bs r} \rangle |}.
\end{equation}

If $\rho(\bs r)<1 $ holds for the current iteration, then $\rho'_{(1)}(\bs r)<1 $ also holds, since $\rho(\bs r)=\rho'_{(1)}(\bs r)$. 

After removing  $j_{(1)}$ from $\Theta_1$ and  $l_{(1)}$ from $ \Theta_2$, the second largest becomes the largest, then we are facing the same problem again with
$\rho'_{(2)}(\bs r)= 
\frac{\max_{\boldsymbol{\psi}}|\langle\boldsymbol{\psi}, \boldsymbol{r}\rangle|}
{\max_{\boldsymbol{\varphi}}|\langle\boldsymbol{\varphi}, \boldsymbol{r}\rangle|}$, 
where $\bs \psi$ ranges over columns in $\Psi$ excluding $\bs \psi_{j_{(1)}}$, and $\bs \varphi$ ranges over columns in $\Phi_\mathrm{opt}$ excluding $\bs \varphi_{j_{(1)}} $.

If condition $\mu_1(l)+ \mu_1(n)<1$ is satisfied, then $ \rho'_{(2)}(\bs r) <1 $ also holds, and so on. Since $ \rho'_{(i)}(\bs r) <1 $ holds for all $i$ until exhausting all indices in $\Theta_1$ and $\Theta_2$, and each term in the summation in \eqref{eq:rho-c-sum} is nonnegative, then $ \rho'_c(\bs r) <1 $ holds. Therefore $\mu_1(l)+ \mu_1(n)<1$ is also sufficient for $\rho_c(\bs r) <1$.  \qed

\vspace{10pt}

Next, we determine the upper bound of $\mu_1(l) + \mu_1(n) $.

$\mu_1(l)+ \mu_1(n) \le \mu (n+l)$ based on the definition of coherence in \eqref{coherence}. Therefore, it is necessary to find the upper bound of $n+l$.

Recall that $n$ is the number of columns in $X$, and $l =\min ( | \Pi |, k-1 )$, where $\Pi$ denotes the index set for the columns in $\Phi_\mathrm{opt}$ that have not been selected by the algorithm so far.

Before the first iteration, $n=k$, and $l=k-1$, $n+l=2k-1$. If $\mu_1(l)+ \mu_1(n)<1$ holds, then $\rho_c(\bs r)<1$ holds, so the greedy algorithm would select as many optimal columns as possible. As long as $|\Pi| \ge c$, it would select $c$ optimal columns. Assume $k \ge c \ge 1$. After the first iteration, $n=k$ and $l=k-c$, therefore $n+l \le 2k-1$. In subsequent iterations, $l$ decreases faster than $n$ increases, thus $n+l$ is not increasing. Therefore the upper bound for 
$n+l$ is
\begin{equation}
\label{eq:n+l}
 n+l \le 2k-1 .
\end{equation}

Lemma \ref{lm-strong-mu} and Eq. \eqref{eq:n+l} lead to Theorem \ref{eq:thm-strong-mu}.

\end{document}